\documentclass[twocolumn, switch]{article} 

\usepackage{preprint}

\usepackage{amsmath, amsthm, amssymb, amsfonts}
\usepackage{algorithm}
\usepackage{algorithmic}

\usepackage[numbers,square]{natbib}

\usepackage[utf8]{inputenc}	
\usepackage[T1]{fontenc}	
\usepackage{xcolor}		
\usepackage[colorlinks = true,
            linkcolor = purple,
            urlcolor  = blue,
            citecolor = cyan,
            anchorcolor = black]{hyperref}	
\usepackage{booktabs} 		
\usepackage{nicefrac}		
\usepackage{microtype}		
\usepackage{lineno}		
\usepackage{float}			

\usepackage{lipsum}		

\usepackage{newfloat}
\DeclareFloatingEnvironment[name={Supplementary Figure}]{suppfigure}
\usepackage{sidecap}
\sidecaptionvpos{figure}{c}

\usepackage{titlesec}
\titlespacing\section{0pt}{12pt plus 3pt minus 3pt}{1pt plus 1pt minus 1pt}
\titlespacing\subsection{0pt}{10pt plus 3pt minus 3pt}{1pt plus 1pt minus 1pt}
\titlespacing\subsubsection{0pt}{8pt plus 3pt minus 3pt}{1pt plus 1pt minus 1pt}

\usepackage{tikz,xcolor,hyperref}

\definecolor{lime}{HTML}{A6CE39}
\DeclareRobustCommand{\orcidicon}{
	\begin{tikzpicture}
	\draw[lime, fill=lime] (0,0)
	circle [radius=0.16]
	node[white] {{\fontfamily{qag}\selectfont \tiny ID}};
	\draw[white, fill=white] (-0.0625,0.095)
	circle [radius=0.007];
	\end{tikzpicture}
	\hspace{-2mm}
}
\foreach \x in {A, ..., Z}{\expandafter\xdef\csname orcid\x\endcsname{\noexpand\href{https://orcid.org/\csname orcidauthor\x\endcsname}
			{\noexpand\orcidicon}}
}

\title{GenTrack2: An Improved Hybrid Approach for Multi-Object Tracking}

\usepackage{xwatermark}
\newwatermark[firstpage,color=gray!90,angle=0,scale=0.28, xpos=0in,ypos=-5in]{*correspondence: \texttt{toanvn@mmmi.sdu.dk}}

\usepackage{authblk}

\author[1]{Toan Van Nguyen\orcidA{}}
\author[2]{Rasmus G. K. Christiansen\orcidB{}}
\author[3]{Dirk Kraft\orcidC{}}
\author[4]{Leon Bodenhagen\orcidD{}}

\affil[*]{SDU Robotics, University of Southern Denmark}

\begin{document}

\twocolumn[ 
  \begin{@twocolumnfalse} 

\maketitle

\begin{abstract}

This paper proposes a visual multi-object tracking method that jointly employs stochastic and deterministic mechanisms to ensure identifier consistency for unknown and time-varying target numbers under nonlinear dynamics. A stochastic particle filter addresses nonlinear dynamics and non-Gaussian noise, with support from particle swarm optimization (PSO) to guide particles toward state distribution modes and mitigate divergence through proposed fitness measures incorporating motion consistency, appearance similarity, and social-interaction cues with neighboring targets. Deterministic association further enforces identifier consistency via a proposed cost matrix incorporating spatial consistency between particles and current detections, detection confidences, and track penalties. Subsequently, a novel scheme is proposed for the smooth updating of target states while preserving their identities, particularly for weak tracks during interactions with other targets and prolonged occlusions. Moreover, velocity regression over past states provides trend-seed velocities, enhancing particle sampling and state updates. The proposed tracker is designed to operate flexibly for both pre-recorded videos and camera live streams, where future frames are unavailable. Experimental results confirm superior performance compared to state-of-the-art trackers. The source-code reference implementations of both the proposed method and compared-trackers are provided on GitHub: \href{https://github.com/SDU-VelKoTek/GenTrack2}{\url{https://github.com/SDU-VelKoTek/GenTrack2}}.
\end{abstract}
\vspace{0.35cm}

  \end{@twocolumnfalse} 
] 



\section{\textbf{Introduction}}
Visual multi-object tracking (MOT) enables a variety of robotic tasks such as object manipulation, human-robot interaction, autonomous navigation in dynamic environments. Despite significant progress, MOT remains challenging due to factors such as object similarity, occlusions, and irregular motion. High similarity between objects can cause identity confusion, while partial or complete occlusions may lead to temporary loss of tracked targets. In addition, irregular or unpredictable object motion further complicates the association of objects across frames. Addressing these challenges is essential for developing robust and accurate MOT systems that can operate effectively in real-world applications.

\subsection{Related Works}
\label{sec:headings}
Multi-object tracking methods are commonly categorized into tracking-by-detection and detection-free paradigms. In settings with variable numbers of targets, tracking-by-detection offers distinct advantages, as frame-wise detections enable the automatic initiation and termination of tracks, and mitigate drift accumulation. These characteristics, reinforced by recent advances in object detection, have established tracking-by-detection as the prevailing approach in state-of-the-art MOT methods. Here, the Kalman filter is widely used for updating motion state and estimating positions, commonly paired with tracklet-detection association. In \cite{c1}, a practical MOT approach, known as SORT, enables real-time tracking but relies heavily on detection quality.  In \cite{c2}, DeepSORT extends SORT with appearance features to reduce identity switches, while ByteTrack \cite{c3} improves association by handling high and low confidence detections separately. In \cite{c4}, BoT-SORT further refines performance with a refined Kalman filter, camera compensation, and motion and appearance cues. In \cite{c5}, OC-SORT handles occlusion noise by estimating virtual trajectories from observations. In \cite{c6}, SMILEtrack builds on ByteTrack and BoT-SORT, integrating an object detector with a Siamese similarity module to enhance appearance similarity compute and matching. In \cite{c7}, ConfTrack adds low-confidence penalization and cascading to a Kalman filter-based framework to handle noisy detections. Notably, Kalman-based trackers assume linear motion and Gaussian noise, yet real-world MOT often involves non-linear dynamics and non-Gaussian noise.

Since the introduction of particle filter-based single-object tracker, numerous innovative approaches have been made to enhance accuracy and robustness. In \cite{c8}, MCMC (Markov Chain Monte Carlo) resampling and kernel-based particle pipelining were proposed to improve computational efficiency. In \cite{c9}, an occlusion-aware particle filter was introduced, integrating color and motion vector features within a patch-based model to manage occlusion effectively. To mitigate degeneracy during importance resampling, particle swarm optimization is employed in \cite{c10}, while the framework further with multi-task correlation filters that jointly exploit interdependencies among features are conducted in \cite{c11}. Chaos theory was incorporated in \cite{c12} through a chaotic particle filter, supported by global motion estimation and color-based refinement. In \cite{c13}, crow search optimization accelerated convergence in the particle filter tracking framework and was paired with an adaptive multi-cue fusion model. In \cite{c14}, a scene-dependent feature fusion is proposed with a partitioned template model, addressing real-world challenges such as shadows and illumination changes. Hybrid quantum particle swarm and adaptive genetic optimization \cite{c15}, as well as a minimax-based sequential Monte Carlo filter \cite{c16}, further improved particle diversity and tracking reliability.

Particle filter–based multi-object tracking faces the challenge of a high-dimensional state space that grows with the number of objects. Independent particle filters for single targets \cite{c17,c18} often fail when similar objects interact, while multi-object particle filters \cite{c19, c20} usually assume a fixed number of targets. Few studies address particle filter with unknown and time-varying target numbers \cite{c21, c22, c23, c24, c25}. The approach in \cite{c21} tackled this but scaled poorly as target numbers increased. To improve the tracking performance, a method in \cite{c22} used a modified Metropolis-Hastings algorithm with add-delete and stay-leave operations, but suffered from duplicate tracks and reduced reliability under frequent target changes. These particle-based methods focus on handling variable target counts, paying less attention to factors like object scale or appearance. In contrast, detection-based approaches \cite{c23, c24, c25} handle target addition and removal more deterministically and emphasize better observation models, though often neglect optimization of the tracking inference.

\subsection{Contributions and Organizations}
The main contributions of this paper include: (1) a hybrid visual multi-object tracking framework that maintains identifier consistency for unknown and time-varying target numbers under nonlinear dynamics by integrating stochastic and deterministic approaches. A novel target update scheme is introduced, incorporating target interactions in state estimation and particle optimization via PSO with tailored fitness measures. Additionally, a more generalized and systematic cost matrix is proposed for matching tracklets to current detections; (2) a velocity regression that leverages historical states to improve particle sampling and state updates, particularly under occlusions or with weak/noisy detectors; and (3) source-code reference implementations of the proposed method and compared trackers to facilitate re-implementation and comparative evaluation.

The remainder of this paper is structured as follows: Section II outlines the methodological background and a visual multi-object tracking framework. Section III evaluates the proposed tracker on human tracking, followed by conclusions in Section IV.

\section{\textbf{Methodology}}
This section begins by formulating the proposed tracking framework, and is then followed by a comprehensive and detailed description of primary processes involved in visual multi-object tracking.

\subsection{Tracking Approach}

Object tracking can be modelled as a nonlinear Bayesian filtering problem, where the posterior state is conditionally estimated from available measurements. In multi-object tracking, targets may enter or leave the scene, requiring a tracker that handles a variable number of targets. Thus, a set of target identifiers is essentially maintained \cite{c21}. The system state is defined as $\{K_t, X_t\}$, with $K_t$ representing the identifiers of \textit{k} current targets, while $X_t$ representing their states. Here, \textit{k} denotes the time-varying number of targets, assuming independent target entries and leaves. In sampling-based methods, the density $P(K_t, X_t | Z_{0:t})$ is approximated with \textit{S} random samples. To overcome challenges like inefficiency in high dimensions, weight degeneracy, proposal sensitivity, and high computational cost, a MCMC approach can be used to formulate the posterior density:        
\begin{equation}
\label{equation1}
\begin{split}
    & P(K_t, X_t | Z_{0:t}) \propto \\ %
    & P(Z_t | K_t, X_t). \sum_{S} P(K_{t}^{s}, X_{t}^{s} | K_{t-1}^{s}, X_{t-1}^{s})       
\end{split} 
\end{equation}

Here, $P(Z_t | K_t, X_t)$  represents the system's observation model, while $P(K_{t}^{s}, X_{t}^{s} | K_{t-1}^{s}, X_{t-1}^{s})$ is estimated using an internal model:

\begin{equation}
\label{equation2}
\begin{split}
    & P(K_t, X_t | K_{t-1}, X_{t-1}) = \\ %
    & P(X_t | K_t, K_{t-1}, X_{t-1}).P(K_t | K_{t-1}, X_{t-1})    
\end{split}
\end{equation}

Sampling in (\ref{equation1}) is defined as $\{K_{t}^{S}, X_{t}^{S}\} \triangleq \sum_{s} P(K_t, X_t | K_{t-1}^{s},X_{t-1}^{s})$, but challenges arise in (\ref{equation2}) because the internal model must not only predict motion but also handle targets entering or leaving. To address this, \cite{c22} splits state $(K_t, X_t)$ into entering $(K_E, X_E)$ and staying $(K_S, X_S)$, applying a Reversible Jump MCMC particle filter with modified Metropolis-Hastings moves. While this approach accommodates varying target counts by adding or deleting tracks during sampling, it is computationally expensive and prone to generating duplicate tracks when particle counts are small, reducing performance in scenarios with frequent target changes. Although stochastic methods can deal with uncertainty and nonlinear dynamics, they yield inconsistent results. Conversely, deterministic methods, such as tracking-by-detection with data association, provide more consistent outputs but rely heavily on the quality of detection and motion models.    

\begin{figure}[!t]
    \centering
    \includegraphics[width=3.3in]{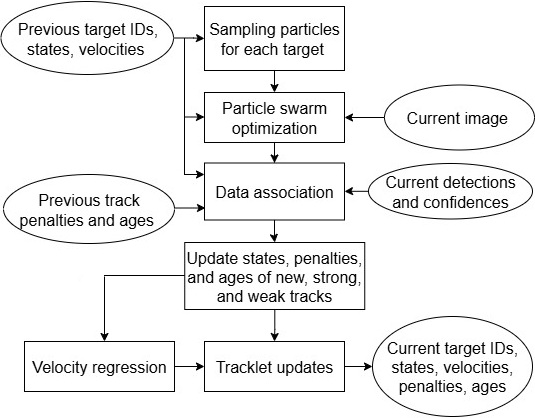}
    \caption{Overview of the proposed method pipeline. The inputs of the tracking system include current image, detections and associated confidences, while previous target IDs, states, velocities, track penalties, and ages are updated internally.}
    \label{fig_1}
\end{figure}

This paper proposes a hybrid multi-object tracking framework combining stochastic methods for non-linear, non-Gaussian motions with deterministic methods to maintain consistent target identities. Sampling follows the approach in (\ref{equation1}) but does not create or remove tracks directly. Each target has a unique birth (track initiation) and death (track removal) determined by tracklet-detection associations via $P(K_t, X_t | Z_t, K_{t-1}^h, X_{t-1}^h, K_{t-1}, X_{t-1})$, with $\{K_t^h, X_t^h\}$ reflects track confidence over time, captured through track history of $\{K_t, X_t\}$, while current detections $\{D_t\}$ and their confidences $\{D_t^{conf}\}$ are integrated into $Z_t$. The deterministic association relies on an object detector, but its weaknesses are mitigated by the stochastic part, allowing robust tracking. Overall, the hybrid approach effectively handles complex dynamics, preserves target identities, and ensures consistent outputs even with frequent changes in target numbers. The pipeline of the proposed method is illustrated as in Figure \ref{fig_1}, while Algorithm \ref{alg:alg1} explains more details of the proposed tracking framework. Here, the multi-object state is given by $\{K_t, X_t\} = \{K_{i,t}, X_{i,t}\}_{i=1}^k$. Target history $\{K_t^h, X_t^h\} = \{K_t, X_t^{pen}, X_t^{age}\}$ includes track penalties $X_t^{pen}$ and ages $X_t^{age}$ of targets up to the current frame $I_t$. Tracklets are now extended with velocity, penalty, and age terms, as $\{K_t, X_t, V_t, X_t^{pen}, X_t^{age}\} = \{K_{i,t}, X_{i,t}, V_{i,t}, X_{i,t}^{pen}, X_{i,t}^{age}\}_{i=1}^k$. For deeper insight into the philosophy of GenTrack, it is encouraged to consult \cite{c31}.

\begin{algorithm}[H]
\caption{The proposed MOT tracking framework.}\label{alg:alg1}
\begin{algorithmic}
\STATE {\textsc{Inputs:}} $\{K_{t-1}, X_{t-1}, V_{t-1}, X_{t-1}^{pen}, X_{t-1}^{age}, I_t,D_t,D_t^{conf}\}$ 
\STATE {\textsc{Particle Sampling:}} 
\STATE \hspace{0.75cm} \textsc{Initial Particles $\{K_t^s, X_t^s, V_t^s\}_{s=1}^S$} 
\STATE \hspace{0.75cm} \textsc{PSO Guided:} 
\STATE \hspace{1.5cm} \textsc{Inputs:} $\{K_t^s, X_t^s, V_t^s\}_{s=1}^S$, and \\ \hspace{2.85cm} $\{K_{t-1}, X_{t-1}, K_t^{nei}, X_t^{nei}, I_t\}$ 
\STATE \hspace{1.5cm} \textsc{Return:} $\{K_{op}^s, X_{op}^s, V_{op}^s\}_{s=1}^S$ and \\ \hspace{3.0cm} $\{K_g^s, X_g^s, f_{g,PSO}^h, X_t^{nei}\}$ 
\STATE {\textsc{Data Association:}}  
\STATE \hspace{0.75cm} \textsc{Inputs:} $\{K_{op}^s,X_{op}^s\}_{s=1}^S$, and \\ \hspace{2.15cm} $\{X_{t-1}^{pen},X_{t-1}^{age},D_t,D_t^{conf}\}$
\STATE\hspace{0.75cm} \textsc{Return:} $\{K_{i,t}^m, X_{i,t}^m,det_{j,t}^m\}_{m=1}^M$ 
\STATE {\textsc{State Updates}} 
\STATE {\textsc{Outputs:}} $\{K_t,X_t,V_t,X_t^{pen},X_t^{age}\}$ 

\vspace{11pt}

\STATE {\textsc{\textbf{Notes:}}} $\{K_{op}^s, X_{op}^s, V_{op}^s\}_{s=1}^S$ denotes optimized particle sets. $\{K_g^s, X_g^s, f_{g, PSO}^h\}$ denotes the global bests and their corresponding history finesses after PSO iterations. $\{K_t^{nei}, X_t^{nei}\}$ consists of current neighbours of targets. $\{K_{i,t}^m, X_{i,t}^m,det_{j,t}^m\}_{m=1}^M$ is a set of M pairs, each consisting of a track and a current detection, $\{K_t^m, X_t^m\} \in \{K_t, X_t\}$, $\{det_t^m\} \in \{D_t\}$.
 \end{algorithmic}
\end{algorithm}

\subsection{Visual MOT Inference Mechanisms}

To capture temporal changes in object appearance, an object’s state at time \textit{t} is defined by its bounding box, as $X_{i,t} = (u_{i,t}, v_{i,t}, w_{i,t}, h_{i,t})$, with center $(u_{i,t}, v_{i,t})$, size $(w_{i,t}, h_{i,t})$, and corresponding velocity $V_{i,t} = (\dot{u}_{i,t}, \dot{v}_{i,t}, \dot{w}_{i,t}, \dot{h}_{i,t})$. Target velocities $V_t = \{V_{1,t}, ..., V_{k,t}\}$ are bounded by $V_t^{max} = \{V_{1,t}^{max}, ..., V_{k,t}^{max}\}$, such that $V_{i,t} \in (-V_{i,t}^{max}, V_{i,t}^{max})$, with $V_{i,t}^{max} = (\dot{u}_{i,t}^{max}, \dot{v}_{i,t}^{max}, \dot{w}_{i,t}^{max}, \dot{h}_{i,t}^{max})$. Subsequently, the three main steps in Algorithm \ref{alg:alg1}, including particle sampling, data association, and state updates, are described in greater detail, highlighting their specific contributions within the context of the visual multi-object tracking. 

\subsubsection{Particle Sampling}

Algorithm \ref{alg:alg1} starts particle sampling from the system’s internal model, generating each particle state $\{K_t^s, X_t^s, V_t^s\}_{s=1}^S$ from either its prior state $\{K_{t-1}^s, X_{t-1}^s, V_{t-1}^s\}_{s=1}^S$ or the previous optimal state of target $\{K_{t-1}, X_{t-1}, V_{t-1}\}$. Since object motion—especially for natural entities—is often unpredictable and lacks explicit models, the random motion model is therefore used in this paper, designed as:      

\begin{equation}
\label{equation3}
\begin{cases}
    V_{i,t} = V_{i,t-1} + \varepsilon_V.U_{V_{i,t}} \\
    X_{i,t} = X_{i,t-1} + \lambda_V.V_{i,t} + \lambda_X.\varepsilon_X.U_{X_{i,t}} \\
\end{cases}
\end{equation}

Here, $\varepsilon_X, \varepsilon_V, \lambda_X$, and $\lambda_V$ control state and velocity exploration and their contributions. $U_X$ and $U_V$ are random position and velocity perturbations, while $U_{X_{i,t}} = (U_{i,t}^u, U_{i,t}^v, U_{i,t}^w, U_{i,t}^h)$ and $U_{V_{i,t}} = (U_{i,t}^{\dot{u}}, U_{i,t}^{\dot{v}}, U_{i,t}^{\dot{w}},  U_{i,t}^{\dot{h}})$ are bounded by $(-U_{X_{i,t}}^{max}, U_{X_{i,t}}^{max})$ and $(-U_{V_{i,t}}^{max}, U_{V_{i,t}}^{max})$ based on the bounding box size in the previous state $\{K_{i,t-1}, X_{i,t-1}, V_{i,t-1}\}$. As a result, motion-based particles can be withdrawn from $\{K_t^s, X_t^s, V_t^s\} \xleftarrow{} \{K_{t-1}^s, X_{t-1}^s, V_{t-1}^s, U_X, U_V\}$, or $\{K_t^s, X_t^s, V_t^s\} \xleftarrow{} \{K_{t-1}, X_{t-1}, V_{t-1}, U_X, U_V\}$.     

Motion-based particles are inherently stochastic and need guidance to reach optimal positions for data association and convergence. It is reminded that the hybrid MOT framework combines stochastic and deterministic methods to handle uncertainty and non-linear systems while maintaining consistent tracking. Deterministic data association handles target birth and death. Rather than full posterior sampling, the focus is on generating optimal particles per target to converge toward their distribution modes. This paper therefore leverages PSO \cite{c26} with a proposed fitness function that refines particles based on three components: history fitness $f_{PSO}^h$ (between current particle and previous optimal target state), exploration fitness $f_{PSO}^p$ (between current particle and its prior PSO update), and social fitness $f_{PSO}^i$ (between particle and its target neighbors), as:      

\begin{equation}
\label{equation4}
    f_{PSO} = \sigma_h.f_{PSO}^h + \sigma_p.f_{PSO}^p + \sigma_i.f_{PSO}^i 
\end{equation}

Here, $\sigma_h, \sigma_p$, and $\sigma_i$ are positive with $\sigma_h + \sigma_p + \sigma_i = 1$. $f_{PSO}^h$ and $f_{PSO}^p$ are measured by the same function $f(\bullet, \bullet)$ but differ by input based on fitness type, as $f_{PSO}^h = f(\{K_{PSO}^s, X_{PSO}^s\}, \{K_{t-1}, X_{t-1}\})$ while $f_{PSO}^p = f(\{K_{PSO}^s, X_{PSO}^s\}, \{K_{PSO}^{s^{'}}, X_{PSO}^{s^{'}}\})$.  

\begin{equation}
\label{equation5}
    f(\bullet, \bullet) = \lambda_s.f_s + \lambda_m.f_m
\end{equation}

In (\ref{equation5}), $f_s = \frac{\langle \vec{X}_i^s, \vec{X}_j \rangle}{|\vec{X}_i^s|.|\vec{X}_j|}$ is cosine similarity between non-negative HoG (Histogram of Oriented Gradients) feature vectors $\vec{X}_i^s$ and $\vec{X}_j$ extracted from bounding boxes $X_i^s = (u_i^s, v_i^s, w_i^s, h_i^s)$ and $X_j = (u_j, v_j, w_j, h_j)$, thus $f_s \in [0, 1]$. Here, $|x|$ denotes the magnitude of the vector \textit{x}, and $\langle a,b \rangle$ denotes the dot product of vectors $\vec{a}$ and $\vec{b}$.  The motion fitness $f_m = 1- \frac{min(|X_i^s -X_j|,d_{o,m})}{d_{o,m}}$, where $d_{o,m}$ is the maximum distance, which adaptively adjusted for each target using its current bounding box size. Positive $ \lambda_s$ and $\lambda_m$ control the contribution of $f_s$ and $f_m $ with $\lambda_s + \lambda_m = 1$.     

Social fitness accounts for interactions among targets in multi-object tracking, which strongly affect performance under occlusion. Such interactions may cause ID switches or track losses, especially under heavy occlusions with many nearby objects. To model this, the neighbours of a target $(K_{i,t}, X_{i,t}) \in (K_t, X_t)$ are first defined through a nearest-neighbour search $\{K_{i,t}, X_{i,t}\}^{nei} = \Psi\langle (K_{i,t}, X_{i,t}), \varepsilon_{nei} \rangle$, where the adaptive threshold $\varepsilon_{nei}$ adjusts per state $X_{i,t}$ based on the size of the bounding box. The resulting neighbour set $\{K_{i,t}, X_{i,t}\}^{nei} = \{K_{j,t}^n, X_{j,t}^n\}_{n=1}^N$ is not used for direct data association (to avoid ambiguity), but instead guides particles to diverge from nearby states, reducing ID switches during occlusion, by contributing to PSO fitness measures and state updates. The social fitness is computed as:       

\begin{equation}
\label{equation6}
\begin{split}
     &f_{PSO}^{i,s} = \frac{\xi_p}{N}.\sum_{j=1}^N \frac{min(|X_{i,t}^s - X_{j,t}|, 2\varepsilon_{nei})}{2\varepsilon_{nei}} \\ %
     &+ \frac{\xi_V}{N}.\sum_{j=1}^N \frac{min(|V_{i,t}^s - V_{j,t}|, V_s^{max})}{V_s^{max}}
\end{split}
\end{equation}

Here, $V_s^{max} = V_{i,t}^{max} + U_{V_{i,t}}^{max}$. Positive $\xi_p$ and $\xi_V$ control state and velocity contributions to social fitness, with $\xi_p + \xi_V = 1$. If $N=0$, then $f_{PSO}^{i,s} = 1$. After PSO, low-fitness particles are discarded or replaced with global bests to guide sampling toward the target distribution mode.     

\subsubsection{Data Association}

The optimized particle sets are used for data association via the Hungarian algorithm \cite{c27}, with a proposed target-oriented cost matrix $C_{mat} \in R^{TxD}$, as:    

\begin{equation}
\label{equation7}
\begin{split}
     C_{i,j} = \lambda_p.\frac{1}{S}. \sum_{s=1}^S C_m^{X_{i,t}^s,det_j}  + \lambda_d.(1-det_j^{conf}) \\ %
     + \lambda_h.X_{i,t-1}^{pen}
\end{split}
\end{equation}

Here, \textit{T} and \textit{D} represent the number of targets and detections. The matching cost $C_{i,j}$ between $X_{i,t}$ and $det_j$ is computed using \textit{S} particles of $X_{i,t}$, track penalty $X_{i, t-1}^{pen} \in [0, 1]$, and confidence $det_j^{conf} \in [0, 1]$ of detection $det_j$, while positive values $\lambda_h, \lambda_d$ and $\lambda_p$ control their contributions to the matching cost, with $\lambda_h + \lambda_d + \lambda_p = 1$. The motion cost $C_m^{X_{i,t}^s,det_j} \in [0, 1]$ between $X_{i,t}^s$ and $det_j$ is defined as:       

\begin{equation}
\label{equation8}
    C_m^{X_{i,t}^s,det_j} = C_{IoU}^{X_{i,t}^s,det_j}.C_d^{X_{i,t}^s,det_j}
\end{equation}

The IoU cost is computed as $C_{IoU}^{X_{i,t}^s, det_j} = 1 - IoU_{det_j}^{X_{i,t}^s}$, while distance cost is $C_d^{X_{i,t}^s,det_j} = \frac{min(|X_{i,t}^s - det_j|, d_{o.d})}{d_{o,d}}$. Here, $det_j = (u_{det}, v_{det}, w_{det}, h_{det})$, with $det_j \in \{D_t\}$, $d_{o,d}$ is the maximum distance derived from $(w_i, h_i)$ and $(w_{det}, h_{det})$, and $IoU_{det_j}^{X_{i,t}^s}$ is intersection of union between $X_{i,t}^s$ and $det_j$. By incorporating detection confidence and track penalties, the proposed method avoids the need for separate sub-matchings as required by some other trackers. 

A set of M pairs $\{K_{i,t}^m, X_{i,t}^m,det_{j,t}^m\}_{m=1}^M$ is then determined via $C_{mat}$, where $(K_{i,t}^m, X_{i,t}^m)$ form strong tracks. Weak tracks are then defined by the residual set $\{K_t^w, X_t^w\} = \{K_t, X_t\} - \{K_{i,t}^m, X_{i,t}^m\}$, while unmatched detections $\{det_t^u, det_t^{u,conf}\} = \{D_t, D_t^{conf}\} - \{det_t^m, det_t^{m, conf}\}$ are used to initialize new tracks $\{K_t^n, X_t^n\}$. Thus, the system state is now $\{K_t, X_t\} = \{(K_t^m, K_t^w, K_t^n), (X_t^m, X_t^w, X_t^n)\}$.

\subsubsection{State Updates}

The states of new, strong, and weak tracks are updated differently. Specifically, $(K_t^m, X_t^m)$ is updated as follows:

\begin{equation}
\label{equation9}
\begin{cases}
    \{K_t^m,X_t^m\} \xleftarrow{} \{det_t^m\} \\
    X_t^{m,pen} = 0 \\
    X_t^{m,age} = 0 \\
\end{cases}
\end{equation}

Here, the state of a strong track is updated as $X_t^m = \frac{1}{2}(X_{t-1}^m + det_t^m)$ if $|X_{t-1}^m - det_t^m| \ge d_o$ where $d_o$ depends on its latest bounding box size; otherwise, it is equal to $det_t^m$.

Besides, new tracks $(K_t^n, X_t^n)$ are initiated when the detection confidence exceeds a threshold, and its state is consistently updated using the detection’s bounding box, as: 

\begin{equation}
\label{equation10}
\begin{cases}
    \{K_t^n,X_t^n\} \xleftarrow{} \{det_t^u,det_t^{u,conf}\} \\
    X_t^{n,pen} = 0 \\
    X_t^{n,age} = 0 \\
\end{cases}
\end{equation}

Weak tracks $(K_t^w, X_t^w)$, lacking matched detections, must be updated carefully to prevent ID switches or loss, using their global bests and neighbours $\{K_g^i, X_g^i, f_{g,PSO}^{i,h}, X_i^{nei}\}_{i=1}^k$. It is noted that if a weak track has no neighours outputed post-PSO, an expanded search $\Psi\langle (K_{i,t}, X_{i,t}), \varepsilon_{nei}^{exp} \rangle$ is conducted by using an expanded neighbour condition $\varepsilon_{nei}^{exp}$. When the neighbour is a strong track, its state follows the matched detection; otherwise, the previous optimal state is used. States of weak tracks are then updated as: 

\begin{equation}
\label{equation11}
\begin{cases}
    \{K_t^w,X_t^w\} \xleftarrow{} \{K_{g}^w,X_{g}^w,f_{g,PSO}^{w,h}, V_t^w, X_w^{nei}\} \\
    X_t^{w,pen} = X_{t-1}^{w,pen} + \zeta.\Delta_t \\
    X_t^{w,age} = X_{t-1}^{w,age} + \zeta.\Delta_t.\partial_{max}\\
    \Delta_t = (1 - e^{\frac{-l^2}{2\sigma^2}}).(1-f_{g, PSO}^{w,h} + \Delta_e)\\
\end{cases}
\end{equation}

Here, $\partial_{max}$ is the maximum age, while $\zeta$ is recovery trust so that $\zeta = sign(\rho_{re} - f_{g,PSO}^{w,h} + \Delta_e)$ if the weak track has neighbours as strong tracks, otherwise $\zeta = 1$. Additionally, $\Delta_e$ is an optional entrance penalty to expedite the removal of tracks inside custom entrance areas, while \textit{l} is the count of consecutive frames the target has no matched detections, $\sigma = \frac{\partial_{max}}{6}$, and $\rho_{re} \in [0, 1]$ is the track recovery threshold. Track penalties and ages satisfy $1 \ge X_t^{w,pen} \ge 0$, and $ \partial_{max} \ge X_t^{w,age} \ge 0$.       

The weak track in the first line of (\ref{equation11}) is updated based on the states of its neighbours. Since all neighbours have already contributed to particle optimization in PSO, the weak track is refined according to their reliability: strong-track neighbours are considered trustworthy and influence the update, whereas untrusworthy neighbours have no effect. Let $\{X_t^b, V_t^b \} \subseteq \{K_t^m, X_t^m, V_t^m\}$ denote the set of trustworthy neighbours, and neighour median $\{\overline{X}_t^b, \overline{V}_t^b \}$, defined as the median of $\{X_t^b, V_t^b \}$. During occlusion, updating bounding box dimensions in the same manner as the center can lead to unrealistic growth or shrinkage over time. Specifically, the predicted size may shrink toward zero or diverge during prolonged occlusions. Therefore, it is advisable to decouple position and size updates. Let ${}^{[u,v]}X_t$ and ${}^{[u,v]}V_t$ denote the bounding box center's position and velocity.  

In the case of $\{X_t^b, V_t^b \} = \emptyset$ or $|{}^{[u,v]}\overline{V}_t^b| < \tau_V$, with $\tau_V$ is a threshold determined by the latest size of weak track, the state will be updated by using its own velocity as:

\begin{equation}
    \label{equation12}
    {}^{[u,v]}X_t^w = {}^{[u,v]}X_{t-1}^w + {}^{[u,v]}V_t^w  
\end{equation}

Otherwise, the state of a weak track will be updated as:

\begin{equation}
\label{equation13}
\begin{cases}
    \delta = \frac{\langle {}^{[u,v]}V_t^w, {}^{[u,v]}\overline{V}_t^b \rangle}{|{}^{[u,v]}V_t^w|.|{}^{[u,v]}\overline{V}_t^b|} \\
    X_t^w = \overline{X}_t^b - \overline{X}_{t-1}^b + X_{t-1}^w, & \text{if } \delta \ge \delta_d   \\
    X_t^w = (1-\sigma_g).X_t^o + \sigma_g.X_g^w, & \text{otherwise} \\
\end{cases}
\end{equation}

Here, $\delta$ represents the cosine similarity between the velocity vectors of the weak track and its neighbour, with $\delta_d$ typically in the range [0.8, 0.95]. Besides, $X_g^w$ and $\sigma_g$ are the post-PSO global best of the weak track and its trust factor on its global best, respectively, while $X_t^o$ is updated same as the obstacle avoidance mechanism, as:  
\begin{equation}
\label{equation14}
\begin{cases}
     {}^{[u,v]}X_t^o = {}^{[u,v]}X_{t-1}^w + {}^{[u,v]}V_t^o\\
     {}^{[u,v]}V_t^o = {}^{[u,v]}V_t^w + \varepsilon_o.\frac{\nabla X_t}{|\nabla X_t|}\\
     \nabla X_t \perp \Delta X_t\\
     \Delta X_t = {}^{[u,v]}X_t^w - {}^{[u,v]}\overline{X}_t^b
\end{cases}
\end{equation}

Here, $\varepsilon_o = \varepsilon_s.\frac{|{}^{[u,v]}V_t^w|}{|\Delta X_t|}.d_{X^w}$, with $d_{X^w}$ denotes the diagonal of the weak track's bounding box, and $\varepsilon_s$ is a small positive controlling the repulsive force from neighbours. Furthermore, to ensure that ${}^{[u,v]}V_t^o$ consistently pushes the weak track away from its neighbours, $\nabla X_t = - \nabla X_t$ if $\langle \nabla X_t, {}^{[u,v]}\overline{V}_t^b \rangle > 0$.  

It is noted that the weak track state is updated only if $|{}^{[u,v]}V_t^w| \ge \tau_V$ to suppress noise. Its size can be finely adjusted with minimal changes and is otherwise updated only when a subsequent detection match occurs. After all state updates, the system tracks penalties and ages are $\{X_t^{pen}\} = \{X_t^{m, pen}, X_t^{w, pen}, X_t^{n, pen}\}$ and $\{X_t^{age}\} = \{X_t^{m, age}, X_t^{w, age}, X_t^{n, age}\}$, respectively. Expired tracks and their data are then removed from tracklets, as follows:  

\begin{equation}
\label{equation15}
\begin{split}
    & \{K_t,X_t,V_t,X_t^{pen},X_t^{age}\} = \\ %
    & \{K_{i,t},X_{i,t},V_{i,t},X_{i,t}^{pen},X_{i,t}^{age} | X_{i,t}^{age} < \partial_{max}\}_{i=1}^k
\end{split}
\end{equation}

Track states have been updated, yet estimating the target velocity for the next frame remains challenging. Random motion models cannot reliably predict target velocity, which critically affects particle generation and state updates in the next frame, especially for weak tracks that lack matched detections during heavy occlusions. While linear motion with a strong enough detector permits simple velocity smoothing as a low-pass or high-pass filter, noisy detections can induce non-linearities even the actual motion of objects is linear. In batch-based trackers, tracklets can be refined using full video detection sequences; however, this approach is limited to pre-recorded videos. In contrast, the proposed tracker is designed to operate flexibly for both pre-recorded videos and camera live streams, where future frames are unavailable. Accordingly, a velocity regression is performed over a sequence of \textit{H} past tracks $\{X_{i,t-H}, ..., X_{i,t}\}$ of each current target $X_{i, t}$ to estimate its trend-based velocity for particle generation and state updates in the next frame. The trend-seed velocity is regressed for each element of state, as:

\begin{equation}
\label{equation16}
{}^dV_{i,t} =
\begin{cases}
    0, &\text{if }Q = 0\\
    \gamma_{\frac{Q+1}{2}}, &\text{if } \frac{Q+1}{2} \in \mathbb{Z} \\
    \frac{1}{2}.(\gamma_{\frac{Q}{2}} + \gamma_{\frac{Q}{2}+1}), &\text{if } \frac{Q}{2} \in \mathbb{Z} \\
\end{cases}
\end{equation}

Here, \textit{d} is the index of each element in the state vector, and $d=1,..,4$ for the track state in this paper. \textit{Q} is the length of ${}^d\Gamma^{\uparrow} = \{{}^d\gamma_1, ..., {}^d\gamma_Q | {}^d\gamma_i \le {}^d\gamma_j, i < j\}$, which is an incrementally sorted list of a slope list ${}^d\Gamma = \{{}^d\gamma_{i,j} = \frac{{}^dX_j - {}^dX_i}{j - i} | 0 \le i < j < H, j \le i + F, |{}^d\gamma_{i,j}| \le \tau\}$, calculated over \textit{H} past states, with a frame window \textit{F} ($0 < F \le H$). The threshold $\tau$ of each target is adaptively adjusted using its current $(w_i, h_i)$.  

\section{\textbf{Evaluations}}

The proposed method is evaluated on the MOT17 dataset \cite{c28}, a widely recognized benchmark for single-camera multi-object tracking. Furthermore, comparisons are made with state-of-the-art trackers, including SORT \cite{c1}, DeepSORT \cite{c2}, ByteTrack \cite{c3}, BoT-SORT \cite{c4}, OC-SORT \cite{c5}, SMILEtrack \cite{c6}, and ConfTrack \cite{c7}, using the same detections and ground truth. In detail, the performance of the trackers was tested on the MOT17-04 sequence, which includes 1050 frames at a resolution 1920×1080 and was recorded at 30 frames per second. Although the dataset is primarily designed for pedestrian tracking, this study tracks all humans in the scene, providing a challenging evaluation with weak and noisy detections. Ground truth was adjusted for this goal and is included with the source code, retaining its original name. A source-code reference implementation of the proposed method is provided with minimal dependencies. Implementations of the compared trackers are also included for fair re-evaluation. All experiments were conducted on a PC with an AMD Ryzen 7 PRO 8840U processor, 16GB RAM, without GPU (Graphics Processing Unit).

\begin{figure}[!t]
    \centering
    \includegraphics[width=3.3in]{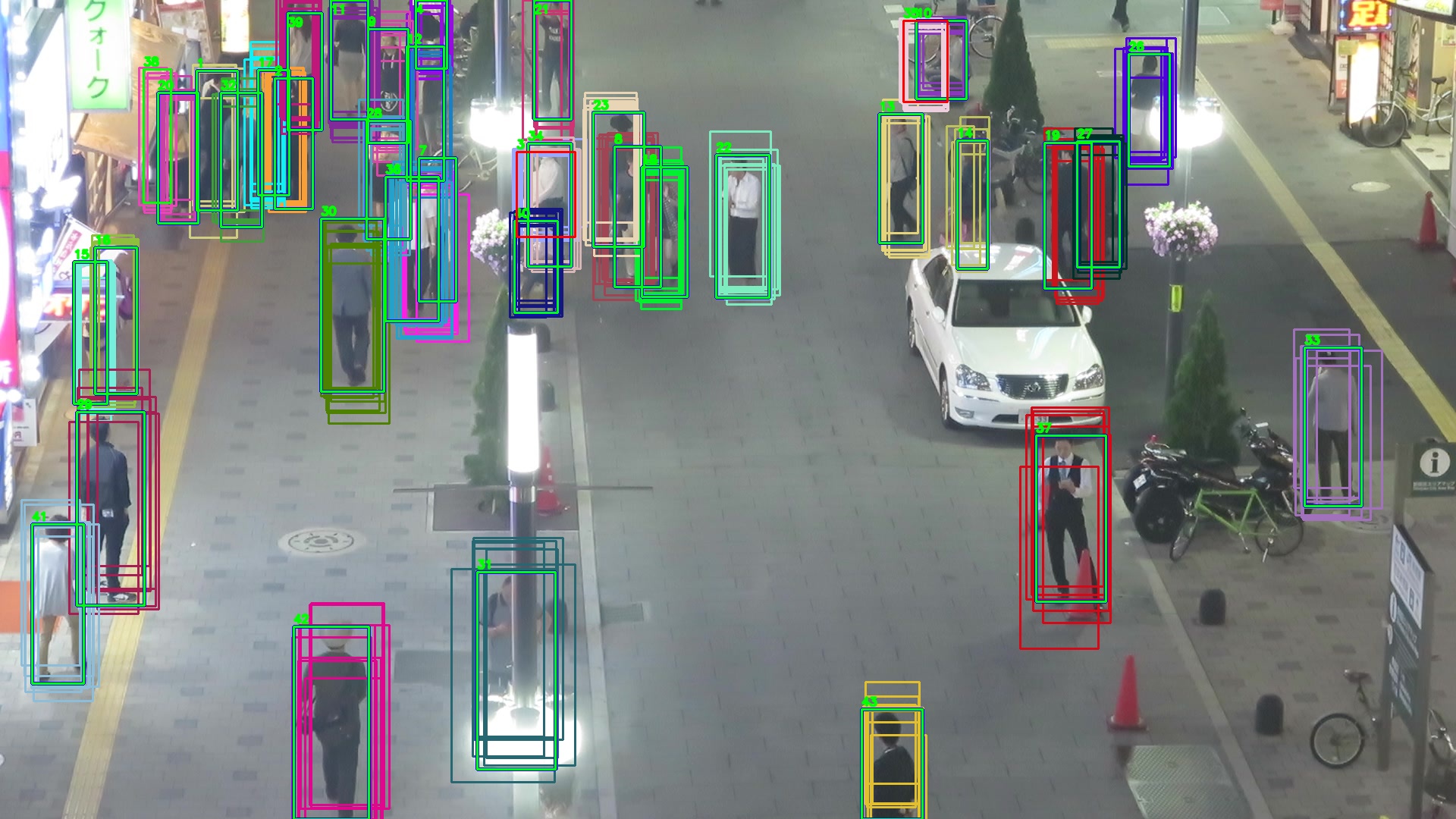}
    \caption{An illustration of visual human tracking, with particle visualization.}
    \label{fig_2}
\end{figure}

\begin{figure}[!t]
    \centering
    \includegraphics[width=3.3in]{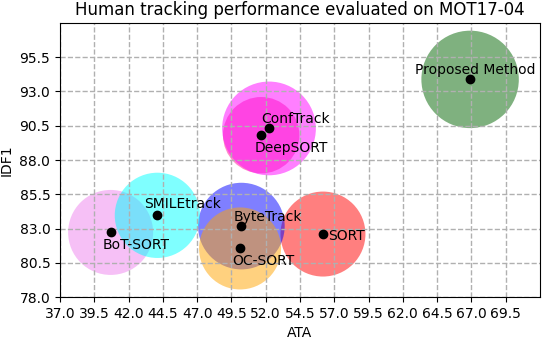}
    \caption{ATA-IDF1-HOTA comparisons of trackers on human tracking.}
    \label{fig_3}
\end{figure}

\begin{figure*}
    \centering
    \includegraphics[width=\textwidth]{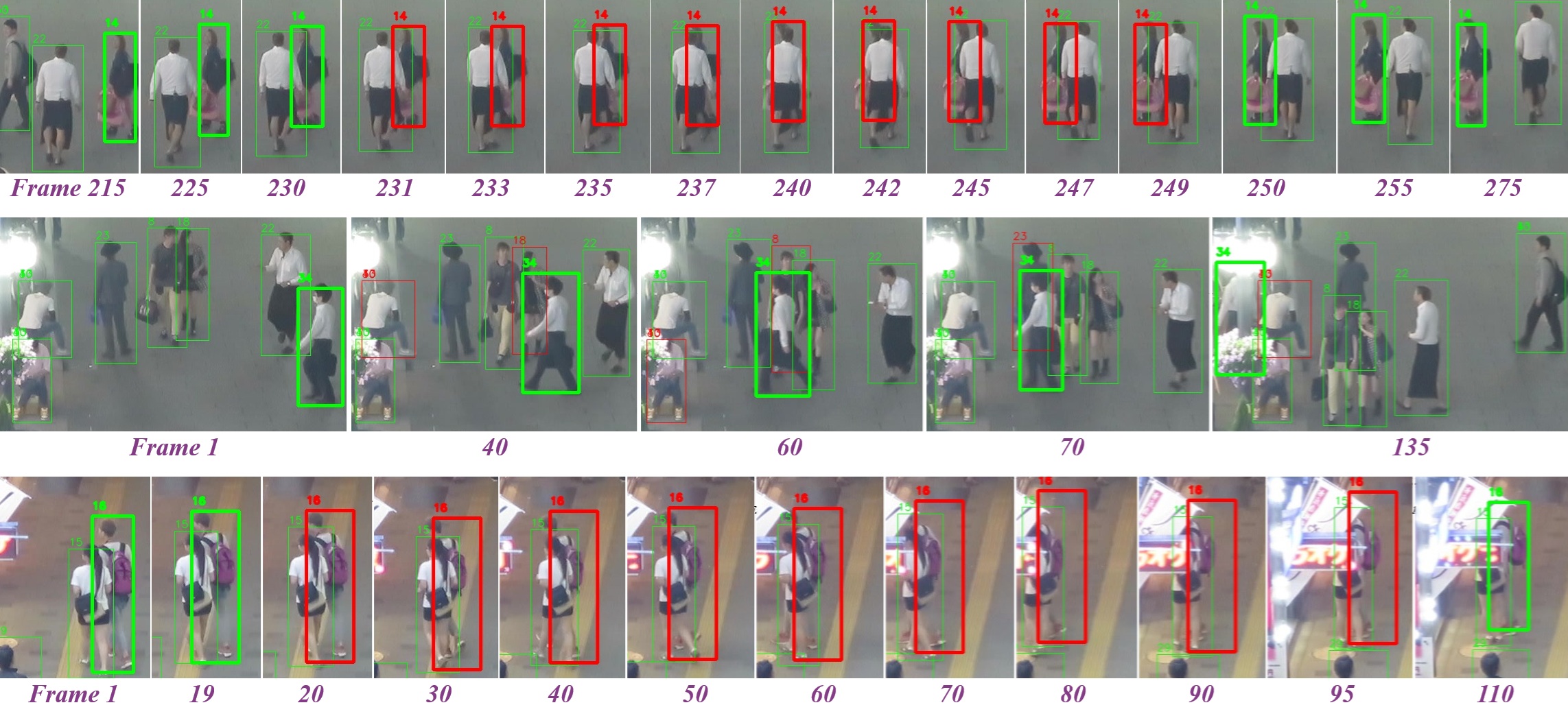}
    \caption{Examples of target interactions during multi-object tracking. Here, green bounding boxes denote strong tracks, red indicate weak tracks, and bold mark observed targets. The first row shows the target ID 14 from frame 215 to 275, with occlusion between frames 231-249. The second row shows the interaction between target ID 34 and neighbours from frame 1 to 135. The last row shows the interaction of target ID 16 from frame 1 to 110, with occlusion between frames 20-95. Target IDs 14 and 34 cross other targets, while target ID 16 moves alongside a neighbour. Target states are updated smoothly, and their IDs are consistently maintained throughout interactions, including during prolonged occlusions.}
    \label{fig_4}
\end{figure*}

Traker performance is assessed using metrics from \cite{c29, c30}, where ATA (Average Tracking Accuracy), IDF1, HOTA (Higher Order Tracking Accuracy), and MOTA (Multiple Object Tracking Accuracy) range from 0 to 100, with higher values indicate better results. By contrast, lower IDSW (ID switches) is preferable. Figure \ref{fig_2} illustrates the scenario used to evaluate tracking performance, with results summarized in Table \ref{tab:table1} and Figure \ref{fig_3}. As shown, the proposed method outperforms existing state-of-the-art trackers across multiple metrics. In particular, ATA and IDSW show significant improvements compared to other methods, while IDF1, HOTA, and MOTA also demonstrate competitive performance, indicating that the proposed method maintains robust performance across different evaluated criteria. Finally, Figure \ref{fig_4} shows several instances of target interactions within the context of multi-object tracking. The states of individual targets are updated in a continuous and coherent manner, and their IDs are consistently maintained even throughout prolonged occlusions.

Insights from implementation reveal that the method is largely insensitive to parameter settings. Particle generation can use either the previous optimal state or prior particles, with the former preferred due to the lack of a motion model. Random motion for all particles may lead to divergence, requiring post-processing. Post-PSO resampling (global or discard) aligns particles with distribution modes, though discard thresholds need care when using few particles. These options are also included in source-code for comparisons. Unlike conventional particle filters, the method achieves strong performance with only 8 particles per target, enabling real-time application. Noise bounds and inter-state motion are based on current bounding box dimensions, and all fitness and cost metrics are normalized to [0, 1] to support systematic analysis and additional optimizations. 

\begin{table}[!t]
    \caption{Human Tracking Performance Evaluated on the MOT17-04 Sequence \cite{c28}\label{tab:table1}}
    \centering
    \scriptsize
    \setlength{\tabcolsep}{5pt}
    \renewcommand{\arraystretch}{1.1}
    \begin{tabular}{|l||c||c||c||c||c|}

    \hline
    {\textbf{Tracker}} & {\textbf{ATA}} & {\textbf{IDF1}} & {\textbf{HOTA}} & {\textbf{MOTA}} & {\textbf{IDSW}}\\
    \hline
    {SORT \cite{c1}} & {56.174} & {82.600} & {61.962} & {69.860} & {103} \\
    \hline
    {DeepSORT \cite{c2}} & {51.668} & {89.808} & {55.701} & {75.574} & {11}\\
    \hline
    {ByteTrack \cite{c3}} & {50.224} & {83.181} & {63.169} & {74.394} & {85}\\
    \hline
    {BoT-SORT \cite{c4}} & {40.690} & {82.718} & {61.970} & {71.417} & {316}\\
    \hline
    {OC-SORT \cite{c5}} & {50.128} & {81.551} & {59.802} & {69.628} & {280}\\
    \hline
    {SMILEtrack \cite{c6}} & {44.101} & {83.965} & {62.172} & {71.484} & {278}\\
    \hline
    {ConfTrack \cite{c7}} & {52.241} & {90.302} & {68.345} & {82.778} & {58}\\
    \hline
    {\textbf{Proposed Method}} & {\textbf{66.903}} & {\textbf{93.878}} & {\textbf{70.044}} & {\textbf{85.717}} & {\textbf{3}}\\
    \hline

\end{tabular}
\end{table}

\section{\textbf{Conclusions}}

This paper proposes a hybrid visual multi-object tracking framework that integrates stochastic particle filter with deterministic association to ensure target identifier consistency under nonlinear dynamics and unknown time-varying numbers of targets. Particle optimization using PSO guides stochastic particles toward their state distribution modes, while a systematic cost matrix enhances data association. Moreover, target interactions are employed both in the PSO fitness measures and in the state updates of unmatched tracks. Furthermore, a velocity regression incorporating a novel scheme improves the smooth updating of target states while preserving their identities, especially for weak tracks during prolonged occlusions. Experiments demonstrate superior performance compared to state-of-the-art trackers. Future research will focus on target interaction models to improve state updates during occlusions and to extract target social behavior patterns.



\begin{thebibliography}{99}

\bibitem{c1} 
A. Bewley et al., ``Simple online and realtime tracking," in 2016 IEEE International Conference on Image Processing (ICIP), Phoenix, AZ, USA, September 2016, pp. 3464-3468.

\bibitem{c2} 
N. Wojke, A. Bewley, and D. Paulus, ``Simple online and realtime tracking with a deep association metric," in 2017 IEEE International Conference on Image Processing (ICIP), Beijing, China, September 2017, pp. 3645-3649.

\bibitem{c3} 
Y. Zhang et al., ``ByteTrack: Multi-object tracking by associating every detection box," In: Avidan, S., Brostow, G., Cisse, M., Farinella, G.M., Hassner, T. (eds) Computer Vision – ECCV 2022. ECCV 2022. Lecture Notes in Computer Science, vol 13682. Springer, Cham, 2022.

\bibitem{c4} 
N. Aharon, R. Orfaig, and B.-Z. Bobrovsky, ``BoT-SORT: Robust associations multi-pedestrian tracking," arXiv:2206.14651v2, 2022.

\bibitem{c5} 
J. Cao et al., ``Observation-Centric SORT: Rethinking SORT for robust multi-object tracking," in 2023 IEEE/CVF Conference on Computer Vision and Pattern Recognition (CVPR), Vancouver, BC, Canada, June 2023, pp. 9686-9696. 

\bibitem{c6} 
Y.-H. Wang et al., ``SMILEtrack: Similarity learning for occlusion-aware multiple object tracking," in Proceedings of the Thirty-Eighth AAAI Conference on Artificial Intelligence, vol. 38, no. 6, pp. 5740-5748, 2024. 

\bibitem{c7}
H. Jung et al., ``ConfTrack: Kalman filter-based multi-person tracking by utilizing confidence score of detection box," in 2024 IEEE/CVF Winter Conference on Applications of Computer Vision (WACV), Waikoloa, HI, USA, January 2024, pp. 6583-6592.

\bibitem{c8}
D. Wang, Q. Zhang, and J. Morris, ``Distributed Markov Chain Monte Carlo kernel based particle filtering for object tracking," Multimedia Tools and Applications, vol. 56, pp. 303-314, 2012.

\bibitem{c9}
S. D. Lin, J.-J. Lin, and C.-Y. Chuang, ``Particle filter with occlusion handling for visual tracking," IET Image Processing, vol. 9, no. 11, pp. 959-968, 2015.

\bibitem{c10}
G. Xia and S. A. Ludwig, ``Object-tracking based on particle filter using particle swarm optimization with density estimation," in 2016 IEEE Congress on Evolutionary Computation (CEC), Vancouver, BC, Canada, July 2016, pp. 4151-4158.

\bibitem{c11}
T. Zhang, C. Xu, and M.-H. Yang, ``Multi-task correlation particle filter for robust object tracking," in 2017 IEEE Conference on Computer Vision and Pattern Analysis (CVPR), Honolulu, HI, USA, July 2018, pp. 4819-4827.

\bibitem{c12}
M. Firouznia et al., ``Chaotic particle filter for visual object tracking," Journal of Visual Communication and Image Representation, vol. 53, pp. 1-12, 2018.

\bibitem{c13}
G. S. Walia et al., ``Robust object tracking with crow search optimized multi-cue particle filter," Pattern Analysis and Applications, vol. 23, pp. 1439-1455, 2020.

\bibitem{c14}
J. Panda and P. K. Nanda, ``Particle filter-based video object tracking using feature fusion in template partitions," The Visual Computer, vol. 39, pp. 2757-2779, 2023.

\bibitem{c15}
X. Yuqi, W. Yongjun, and Y. Fan, ``A scale adaptive generative target tracking method based on modified particle filter," Multimedia Tools and Applications, vol. 82, pp. 31329-31349, 2023.

\bibitem{c16}
J. Lim, J.-Y. Park, and H.-M. Park, ``Minimax Monte Carlo object tracking," The Visual Computer, vol. 39, pp. 1853-1868, 2023.

\bibitem{c17} 
C. Yang, R. Duraiswami, and L. Davis, ``Fast multiple object tracking via a hierarchical particle filter," in Proceedings of Tenth IEEE International Conference on Computer Vision (ICCV), Beijing, China, October 2005, pp. 212-219.

\bibitem{c18} 
W. Hu et al., ``Single and multiple object tracking using log-Euclidean Riemannian subspace and block-division appearance model," IEEE Transactions on Pattern Analysis and Machine Intelligence, vol. 34, no. 12, pp. 2420-2440, 2012.

\bibitem{c19} 
Y. Jin and F. Mokhtarian, ``Variational particle filter for multi-object tracking," in 2007 IEEE 11th International Conference on Computer Vision (ICCV), Rio de Janeiro, Brazil, October 2007, pp. 1-8.

\bibitem{c20} 
R. Hess and A. Fern, ``Discriminatively trained particle filters for complex multi-object tracking," in 2009 IEEE Conference on Computer Vision and Pattern Recognition (CVPR), Miami, FL, USA, June 2009, pp. 240-247.

\bibitem{c21} 
M. Isard and J. MacCormick, ``BraMBLe: A Bayesian multiple-blob tracker," in Proceedings Eighth IEEE International Conference on Computer Vision (ICCV 2001), Vancouver, BC, Canada, July 2001, pp. 34-41.

\bibitem{c22} 
Z. Khan, T. Balch, and F. Dellaert, ``MCMC-based particle filter for tracking a variable number of interacting targets," IEEE Transactions on Pattern Analysis and Machine Intelligence, vol. 27, no. 11, pp. 1805-1819, 2005.

\bibitem{c23} 
M. D. Breitenstein et al., ``Robust tracking-by-detection using a detector confidence particle filter," in 2009 IEEE 12th International Conference on Computer Vision (ICCV), Kyoto, Japan, September-October 2009, pp. 1515-1522.

\bibitem{c24} 
M. Yang et al., ``Detection driven adaptive multi-cue integration for multiple human tracking," in 2009 IEEE 12th International Conference on Computer Vision (ICCV), Kyoto, Japan, September-October 2009, pp. 1554-1561.

\bibitem{c25} 
K. Okuma et al., ``A boosted particle filter: multitarget detection and tracking," in Proceedings of European Conference on Computer Vision (ECCV), Florence, Italy, May 2024, pp. 28-39.

\bibitem{c31}
Toan Van Nguyen et al., "GenTrack: A New Generation of Multi-Object Tracking", arXiv, 2025. Available at: \href{https://doi.org/10.48550/arXiv.2510.24399}{\url{https://doi.org/10.48550/arXiv.2510.24399}}

\bibitem{c26} 
J. Kennedy and R. Eberhart, ``Particle swarm optimization," in Proceedings of IEEE International Conference on Neural Networks, Perth, Australia, November 1995, pp. 1942-1948.

\bibitem{c27}
H. W. Kuln, ``The Hungarian method for the assignment problem," Naval Research Logistics Quarterly, vol. 2, no. 1-2, pp. 83-97, 1955.

\bibitem{c28}
P. Dendorfer et al., ``MOTChallenge: A benchmark for single-camera multiple target tracking," International Journal of Computer Vision, vol. 129, pp. 845-881, 2021.

\bibitem{c29}
V. Manohar et al., ``Performance evaluation of object detection and tracking in video," in: Narayanan, P.J., Nayar, S.K., Shum, HY. (eds) Computer Vision – ACCV 2006. ACCV 2006. Lecture Notes in Computer Science, vol. 3852, pp. 151-161. Springer, Berlin, Heidelberg.

\bibitem{c30}
J. Luiten et al., ``HOTA: A higher order metric for evaluating multi-object tracking," International Journal of Computer Vision, vol. 129, pp. 548-578, 2021.

\end{thebibliography}

\section*{Acknowledgment}

This work was a part of VelKoTek project, funded by The Ministry of Food, Agriculture and Fisheries of Denmark via the Green Development and Demonstration Program (GUDP).


\end{document}